\def\year{2020}\relax
\documentclass[letterpaper]{article} % DO NOT CHANGE THIS
\usepackage{aaai20}  % DO NOT CHANGE THIS
\usepackage{times}  % DO NOT CHANGE THIS
\usepackage{helvet} % DO NOT CHANGE THIS
\usepackage{courier}  % DO NOT CHANGE THIS
\usepackage[hyphens]{url}  % DO NOT CHANGE THIS
\usepackage{graphicx} % DO NOT CHANGE THIS
\usepackage{graphicx}  % DO NOT CHANGE THIS
\usepackage{xcolor}  % DO NOT CHANGE THIS
\nocopyright

\usepackage{amssymb}
\usepackage{amsmath}
\usepackage{amsfonts}
\usepackage[ruled,vlined]{algorithm2e}

\SetCommentSty{mycommfont}

\title{Explainable Composition of Aggregated Assistants}
\author{
Sarath Sreedharan$^1$\thanks{Work done as an intern at IBM Research AI, Cambridge (USA) during the summer of 2020.} $\cdot$
Tathagata Chakraborti$^2$ $\cdot$
Yara Rizk$^2$ $\cdot$
Yasaman Khazaeni$^2$\\[1ex]
$^1$Arizona State University, Tempe AZ USA 85281\\
$^2$IBM Research AI, Cambridge MA USA 02142\\[1ex]
ssreedh3@asu.edu, \{ tchakra2, yara.rizk \} @ibm.com, yasaman.khazaeni@us.ibm.com
}

\begin{document}

\maketitle

\begin{abstract}

A new design of an AI assistant that has become 
increasingly popular is that of an ``aggregated assistant'' --
realized as an orchestrated composition of 
several individual skills or agents that can each
perform atomic tasks. 
In this paper, we will talk about the role of planning
in the automated composition of such assistants
and explore how concepts in automated planning can help
to establish transparency of the inner workings of
the assistant to the end-user.

\end{abstract}

\noindent Conversational assistants
such as Siri, Google Assistant, and Alexa 
have found increased user adoption 
over the last decade and are adept at 
performing tasks like 
setting a reminder or an alarm, putting 
in an order online, control a smart 
device, and so on. 
However, the capability of such 
assistants remain quite limited to 
episodic tasks that mostly involve
a single step and do not require
maintaining and propagating state
information across multiple steps.

A key hurdle in the design of more 
sophisticate assistants is the complexity
of the programming paradigm -- at the 
end of the day, end-users and developers
who are not necessarily subject matter 
experts have to be able to build and 
maintain these assistants.
A particular architecture that has 
emerged recently to address this issue
is that of an "aggregated assistant" 
where the assistant is build out 
of individual components called skills. 
Skills are the unit of automation 
and they perform atomic tasks that 
can be composed together to 
build the assistant capable of 
performing more complex tasks.
Prominent examples of this 
include IBM Watson Assistant 
Skills\footnote{IBM Watson Assistant Skills \url{https://ibm.co/33f58Hc}}
and Amazon Alexa Skills\footnote{Amazon Alexa Skills: \url{https://amzn.to/35xK2Xv}}.

This setup is not particularly confined to 
personal assistants either. 
An increasingly popular use of assistants is 
in enterprise applications. 
Here also, examples of aggregated 
assistant can be seen in offerings from
Blue Prism\footnote{Blue Prism Digital
Exchange: \url{https://bit.ly/2Ztzdla}},
Automation Anywhere\footnote{Automation Anywhere Bot 
Store: \url{https://bit.ly/33hcr12}}, and others.
These individual skills belong to the class of Robotic
Process Automation (RPA) tools that automate simple
repetitive tasks in a business process. 

With recent advances in AI and conversational assistants,
the scope of RPAs has also been evolving to take on 
more complex tasks, as we outline in \cite{bpm-rpa-forum}.
From the point of view of the planning community, 
this poses an interesting challenge: one of composing
skills in a goal-oriented fashion to automatically
realize assistants that can achieve longer-term 
goal-directed behavior.
In \cite{d3ba} we demonstrated one such possibility 
of optimizing an existing workflow or business processes 
by composing it with skills to maximize automation 
and minimize case worker load. 
We showed how this design can be done offline 
by adopting a non-deterministic planning substrate.
This design choice does not, however, readily 
lend itself to the automated composition of aggregate
assistant -- the composed assistant is going to 
evolve rapidly based on user interactions and it does 
not seem reasonable to model all possibilities over 
all goals up front. 

\begin{figure}
\centering
\includegraphics[width=\columnwidth]{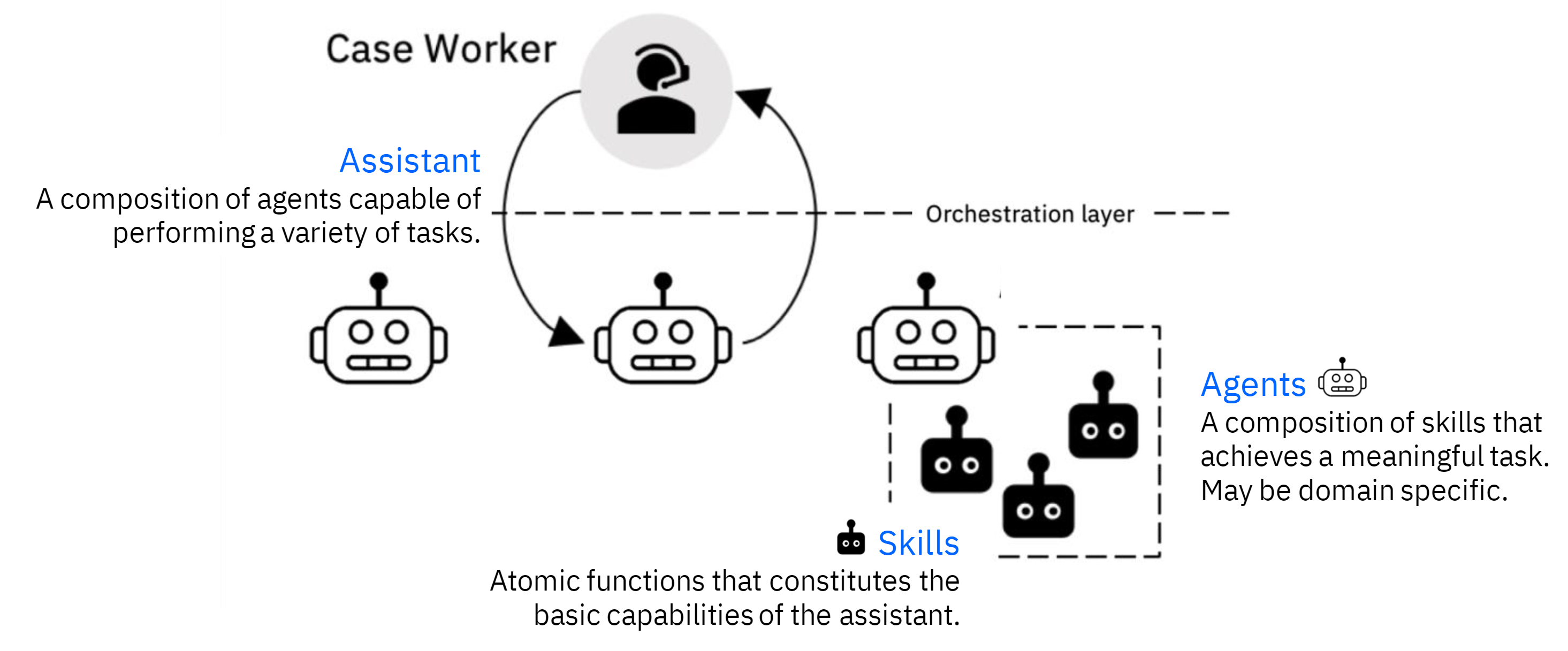}
\caption{
Simplified architecture diagram of Verdi \cite{Rizk2020AUnified}
illustrating the assistant-agent-skill hierarchy.
}
\label{fig:arch}
\end{figure}

Instead, in this paper, we will describe how 
we can use a rapid planning and re-planning loop 
to compose assistants on the fly. 
The first part of the paper will be outlining 
the implementation of this aggregated assistant 
in the form of a continuously evolving planning problem.
One interesting outcome of this design is that 
a lot of the backend processes that affect the 
user interaction do not get manifested externally. 
In the second part of the paper, we will 
explore how concepts of causal chains and landmarks
in automated planning will help us navigate 
transparency issues in aggregate assistants
composed on the fly.

\section{Aggregated Assistants}
\label{definitions}

The particular aggregated assistant we will focus on
is Verdi, as introduced in \cite{Rizk2020AUnified} -- architecturally, 
this subsumes all the examples above. 
Figure \ref{fig:arch} provides a simple view of the system. 
An assistant in Verdi interfaces with the 
end-user (for a personal assistant)
or case worker (for a business process assistant) 
in the form of an orchestrated set of agents, which
are in turn composed of a set of skills.

\subsubsection{Events}

The assistant is set up to handle and respond to 
different kinds of phenomenon in its environment -- 
e.g. a text from the end-user, an alert 
from a service it is monitoring, a data object
or a pointer to a business process, and so on.
In general, we will call them ``events'' $\mathcal{E}$. 
In the examples in this paper, unless otherwise mentioned,
we will deal with events involving user utterances only. 

\subsubsection{Facts}

The assistant also has access to a shared 
memory or knowledge base which stores global
information known to the assistant -- this is 
referred to as the Long Term Memory or LTM in Verdi. 
This information can be accessed by different 
agents and skills inside the assistant.
We will refer to a variable in the LTM as a fact $F$ 
with value $f$. 
For the purposes of this paper, this abstraction will 
be enough and we will not go into the details of 
the architectural implementation of the LTM.
We will also assume that all variables in the LTM
share the same vocabulary (more on this later). 

\subsubsection{Skills}

A skill, as we mentioned before, is an atomic function 
that performs a specific task or transform on data. 
Each skill $\phi$ is defined in terms of the tuple:

\begin{itemize}
\item[] Skill $\phi = \langle [[i, \{o\}_i], \ldots] , \delta \rangle$  
\begin{itemize}
\item[] where $i \subseteq F$ and $o \subseteq F$ 
are the inputs and outputs of 
the skills respectively, and $delta$ is the number of times
the skill can be called in the lifespan of one user session. 
Notice that the outcome is a set: each member $o_i$ is a possible
outcome corresponding to the input $i$, and there are 
multiple such pairs of inputs and corresponding output sets
for each skill.
\end{itemize}
\end{itemize}

For example, a skill that submits a credit card application
for the user would need as an input the necessary
user information and as output will either produce 
a successful or failed application status
or may require further checks. It can also be pinged 
with a different input such as an application number
to retrieve the status of the application.

\subsubsection{Agents}

An agent can then be defined as the tuple: 

\begin{itemize}
\item[] Agent $\Phi = \langle \mathcal{L}, \{\phi\}_i \rangle$  
\begin{itemize}
\item[] where $\phi_i$ is its constituent skills 
and $\mathcal{L}$ is the logic (e.g. a piece of code) 
that binds the skills together. 
\end{itemize}
\end{itemize}

Every agent defines two functions:

\begin{itemize}
% \item[] $\Phi\textit{.preview}\ : \ \Phi \times \mathcal{E} \mapsto \mathbb{R}$
\item[] $\Phi\textit{.preview}\ : \ \Phi \times \mathcal{E} \mapsto \mathbb{E}[\mathcal{E}]$
\begin{itemize}
\item[]
This is the preview function ($\Phi.p$ in short) that provides an estimate
of what will happen if an agent is pinged with an event.
The output of the preview is an expectation on the actual output.
In this paper this expectation will take the form of a probability 
that the agent can do something useful with the input event, thus: 
$\Phi\textit{.preview}\ : \ \Phi \times \mathcal{E} \mapsto [0, 1]$.
\item[]
Note that an agent may or may not have a preview. For example,
if the evaluation requires a state change then an agent 
may not provide a preview: e.g. 
calling a credit score service multiple times can be 
harmful.
\end{itemize}
\item[] $\Phi\textit{.execute}\ : \ \Phi \times \mathcal{E} \mapsto \mathcal{E}$
\item[]
This is the execute function ($\Phi.e$ in short) 
that actually calls an agent to consume an event. 
Every agent must have an execute function. 
\end{itemize}

\subsubsection{Assistant}

The assistant can then be thought of as a mapping from an event and 
a set of agents to a new event:

\begin{itemize}
\item[] Assistant $\mathbb{A}\ :\ \mathcal{E} \times \{ \Phi_i \} \mapsto \mathcal{E}$
\end{itemize}

The exact nature of this mapping determines the 
behavior of the assistant -- this is the orchestration problem.

\begin{algorithm}[tbp!]
\SetAlgoLined
\While{$\mathcal{E}$}{
\vspace{5pt}
$\{\Phi\}_i \times \mathcal{E} \mapsto \Phi_{eval}, \Phi_{eval} \subseteq \{\Phi\}_i$\\
\tcc{subset of agents to preview}
\vspace{5pt}
\ForEach{$\Phi \in \Phi_{eval}$}{
\eIf{$\exists \Phi\textit{.preview}$}{
$\mathbb{E}_\Phi[\mathcal{E}] \leftarrow \Phi\textit{.preview}(\mathcal{E})$\\
\tcc{compute expected outcomes}
}{
$\mathbb{E}_\Phi[\mathcal{E}] \leftarrow \emptyset$
}
}
\vspace{5pt}
$\{\Phi\}_i \times \{\mathbb{E}_\Phi[\mathcal{E}]\}_i \times \mathcal{E} \mapsto \Phi_{exe}, \Phi_{exe} \subseteq \{\Phi\}_i$\\
\tcc{subset of agents to execute}
\vspace{5pt}
\ForEach{$\Phi \in \Phi_{exe}$}{
\Return $\Phi\textit{.execute}(\mathcal{E})$
}
}
\caption{Verdi Flow of Control}
\label{algo}
\end{algorithm}

\section{Orchestration of an Aggregated Assistant}

In this section, we will introduce how different orchestration 
patterns can be developed to model different types of assistants
and describe the role of planning in it.
Algorithm \ref{algo} provides an overview of the
flow of control in an aggregated assistant, between
the orchestrator and its agents. 

\subsubsection{Apriori and Posterior Patterns}

Following the flow of control laid out in Algorithm \ref{algo} 
there are two general strategies for orchestration:

\begin{itemize}
\item[-] {\bf apriori}
where the orchestrator acts as a filter and decides which agents to invoke 
a preview on, based on $\mathcal{E}$; and
\item[-] {\bf posterior}
where all the agents receive a broadcast and respond with their previews, 
and let the orchestrator pick the best response to execute.
\end{itemize}

These two strategies are not mutually exclusive.
The apriori option is likely to have a smaller system footprint, 
but involves a single bottleneck based on the accuracy of the selector  
which determines which agents to preview. 
The posterior option -- despite increased latency and computational load -- 
keeps the agent design less dependent on the orchestration layer 
as long as the confidences are calibrated.\footnote{This 
can be achieved by learning from agent previews over time
and gradually adapting a normalizer or a more sophisticated selection strategy 
over the confidences self-reported by the agents,
e.g. using contextual bandits \cite{sohrabi2010customizing}. 
The ability to learn 
such patterns can also be used to eventually realize a healthy mix 
of apriori and posterior orchestration strategies.
}

\subsection{The S3 Orchestrator} 
\label{sec:s3}

The S3-orchestrator \cite{Rizk2020AUnified} is a 
stateless, posterior orchestration algorithm 
that consists of the following three stages: Scoring, Selecting and Sequencing. 
% Posterior implies that the orchestrator evaluates the responses of agents to determine what agent(s) to handle an event. 
Hence, the first step is to broadcast the event to all agents and solicit a 
preview of their actions: {\em this preview must not cause any side-effects on the state of the world in case the agent is not selected}. 
Once the orchestrator obtains the agents' responses, 
it scores these responses using an appropriate scoring function 
so that the agents can be compared fairly (this is where the 
normalization, as mentioned above, can happen, for example). 
After computing the scores, 
a selector evaluates them and picks one or more agents 
based on its selection criteria. 
One example is simply picking the agents that have the maximum score if greater than a threshold (referred to as \textit{Top-K} selector). 
Finally, if more than one agent is selected for execution, 
a sequencer determines the order in which agents must 
execute to properly handle the event. 
Algorithm \ref{s3} details the process.
Since agents' executions may be dependent on each other, 
the order is important. Currently this is handled through
simple heuristics and rules coded directly into the Sequencer.
We will discuss later how planning and XAIP has a role
to play here as well. 

\subsection{Stateful Orchestration using Planning}

The S3-orchestrator does not maintain state and does not 
make any attempt to reason about the sequence of agents 
or skills being composed. 
In the following, we will go into details of 
a planner-based orchestrator that 
can compose agents or skills on the fly 
to achieve user goals. 
In comparison to the S3, this means we no longer
need a selector-sequencer pattern: the planner
decides which agents or skills to select and 
how to sequence them. 
In order to be able to do this effectively, 
we need to have the following 
external components. 

\subsubsection{Specification.} 
The ability to compose agents on the fly requires
a specification of agents in terms of (at least) 
their inputs and outputs.\footnote{Of course,
this can be extended to include additional 
rules relevant at the orchestration layer, such
as ordering constraints \cite{allen1983maintaining} to model 
preferred sequences: e.g. {\em ``make sure 
to start with the chit-chat agent
before anything else''}. }
This is somewhat similar to applications of 
planning \cite{sohrabi2010customizing,carman2003web} 
in the web service composition 
domain \cite{srivastava2003web}.
A necessary prerequisite to be able
to use these variables from the shared memory
across different agents
and skills (developed independently and from
different sources) is that they share some
vocabulary. This needs to be enforced on the 
specification through the use of some
schema or ontology to ensure consistency up front,
or by fuzzy matching variable names on the fly
during execution. 
Existing works in using planning for web
service composition have explored advanced technique on this 
topic \cite{hoffmann2007web,2009-Weber-PhD,hepp2005semantic,sirin2004htn}.
As we mentioned before, for this paper,
we assumed that the skill and agent specifications
share the same vocabulary.

\begin{algorithm}[tbp!]
\SetAlgoLined
\DontPrintSemicolon
\SetKwFunction{FScorer}{Scorer}
\SetKwFunction{FSelector}{Selector}
\SetKwFunction{FSequencer}{Sequencer}
\SetKwProg{Fn}{Function}{:}{}
\While{$\mathcal{E}$}{
\ForEach{$\Phi \in \{\Phi\}_i$}{
\eIf{$\exists \Phi\textit{.preview}$}{
$P(\Phi) \leftarrow \Phi\textit{.preview}(\mathcal{E})$
\tcp*{confidence}
$S(\Phi) \leftarrow \FScorer(P(\Phi))$
}{
$S(\Phi) \leftarrow 0$
}
}
\vspace{5pt}
$\Phi_{exe} \leftarrow \FSelector(\{S(\Phi)\}_i), \Phi_{exe} \subseteq \{\Phi\}_i$\;
$\langle\Phi\rangle \leftarrow \FSequencer(\Phi_{exe})$\;
\vspace{5pt}
\ForEach{$\Phi \in \langle\Phi\rangle$}{
\Return $\Phi\textit{.execute}(\mathcal{E})$
}
}
\vspace{5pt}
\SetKwProg{Pn}{Function}{:}{\KwRet}
\Pn{\FScorer{$P(\Phi)$}}{
\Return $P(\Phi)$
\tcp*{default}
}
\vspace{5pt}
\SetKwProg{Pn}{Function}{:}{\KwRet}
\Pn{\FSelector{$\{S(\Phi)\}_i$}}{
\Return $\Phi_{ext} = \{ \Phi \ | \ S(\Phi) \geq \delta  \}$ \;
such that: $|\Phi_{ext}| \leq k$ and\\
$\not\exists \Phi \in  \Phi_{ext},  \Phi' \not\in \Phi_{ext}$ with $S(\Phi) > S(\Phi')$\\
\tcc{select top-$k$ above threshold $\delta$}
}
\vspace{5pt}
\SetKwProg{Pn}{Function}{:}{\KwRet}
\Pn{\FSequencer{$\{\Phi\}_i$}}{
\Return {\tt List($\{\Phi\}_i$)}
\tcp*{default}
}
\caption{S3-Orchestrator}
\label{s3}
\end{algorithm}

\begin{algorithm}[tbp!]
\SetAlgoLined
\DontPrintSemicolon
\SetKwFunction{FGoal}{Derive-Goal}
\SetKwFunction{FPlanner}{Planner}
\SetKwProg{Fn}{Function}{:}{}
\While{$\mathcal{E}$}{
$\mathcal{G} \leftarrow \FGoal(\mathcal{E})$\;
$\langle\Phi\rangle \leftarrow \FPlanner(\{\Phi\}_i, \mathcal{G}, \mathcal{E})$\;
\vspace{5pt}
\ForEach{$\Phi \in \langle\Phi\rangle$}{
\Return $\Phi\textit{.execute}(\mathcal{E})$
}
}
\caption{Planner-Based Orchestration}
\label{planner}
\end{algorithm}

\subsubsection{Goal-reasoning engine.} 
Finally, the composition at the orchestration layer 
requires a goal: the planner-based orchestrator 
produces goal-directed sequences of agents and skills.
This goal is something that is derived from 
events and is usually related to something the user 
is trying to achieve. However, this may not always
be the case. We will go into further details 
of this in the next section.

\section{Stateful Orchestration using Planning}

We will now go into the details of implementation of 
the planner-based orchestrator, and then a detailed
example of it in action. This will help us better 
understand the XAIP issues engenderd by automated
orchestration of an aggregated assistant 
and the details of the proposed solution.

\subsection{Agent and skill specification}
\begin{figure}[!th]
\centering
\includegraphics[width=0.9\columnwidth]{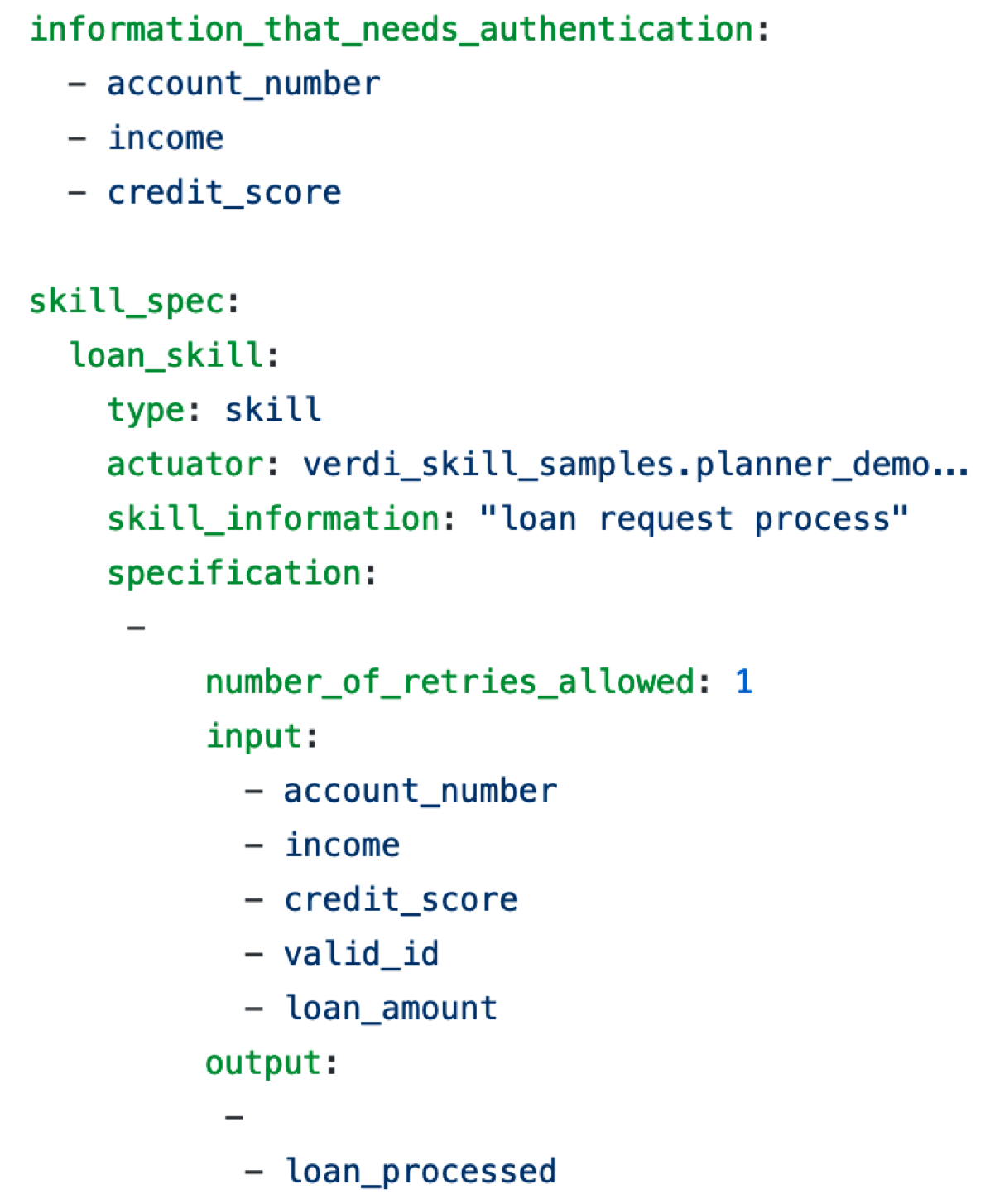}
\caption{A sample specification of a skill.}
\label{fig:spec}
\end{figure}

The central requirement to start converting the agent sequencing problem 
into a planning problem is some form of a model of 
how the skills and agents operate. 
We want to keep the specification as 
simple as possible and move as much of the specifics of 
orchestration patterns to the execution component 
as possible (where we can leverage the evaluate option
whenever possible). 
This not only lets us alleviate some of the modeling overhead 
from developers but also allows us to keep the planning layer 
light thus allowing us to perform rapid execution and replanning.  
% The agent specification file itself would contain a set of general information relevant to the entire agent configuration, a set of specifications for inbuilt Verdi skills that the agent could use, and specifications for the external skills that may be developed by independent developers and may live in a common skill catalog. We require that every element of these specifications are synced via an ontology that could potentially include information correlating the various elements of the specifications.
A skill or agent specification (Figure \ref{fig:spec})
contains the following information:

\begin{itemize}
\item[1.] 
The function endpoint of the skill or agent.
\item[2.] 
A user understandable description of what it does
(to be used in generating explanatory messages).
\item[3.] 
An upper limit on the number of times the assistant can retry 
the same skill or agent to get a desired outcome.\footnote{This is 
important since you may not want the assistant to, for example, 
ping a credit score retrieval skill more than once and inadvertently 
reduce the credit score of the user, or you may not want
the assistant to get stuck at a disambiguation question to
fill a slot when it cannot detect the utterance multiple times.}
\item[4.] 
And finally, an (approximate) specification of functionality
(as described in Section \ref{definitions}). 
as a set of pairs of tuples of input and possible output pairs
that represents roughly the various operational modes of the skills. 
\end{itemize}

For example, a visualization skill could take raw data and generate 
the plot that it believes best represents that data or it could take 
the data along with the plot type to create the specified plot. 
Each pair consists of a set of inputs the skill or agent can take and 
specifies a set of possible outputs that could be generated by the 
skills for those inputs. 
The possible output set consists of mutually exclusive sets of 
output that may be generated. 
These can be thought of as the non-deterministic effects of the skills, 
though in the next section we will discuss how these scenarios diverge 
from traditional non-deterministic planning settings. 
% Each of the input and output elements is assumed to have 
% a corresponding entry in the ontology. 
% For now we will assume that any specific requirements on the values of these items will be reflected by distinct elements in the ontology. For example, if there are two different loan application skills, one for lower loan amounts and another that is meant for higher loan amounts, then we will assume the input for the second skill will be denoted by the element high\_loan\_amount (which may be a subclass of the element loan\_amount). Though going forward we could also allow authors to specify constraints directly, which could be then be used by the planner.
Special inbuilt skills (we will see some examples of these later 
in Section \ref{example})
in Verdi also get an entry in the specification 
with a similar structure to above, but we do not require them to refer 
to elements in the ontology. They may use keywords that may map to specific 
compilations when we build the planning model out of the specification. 
% The agent level specification entry could specify global constraints. 
% For example, one could assign specific properties for certain information 
% items at an agent level and use such properties to specify preferences 
% over certain sequences.

\begin{figure}[!th]
\centering
\includegraphics[width=\columnwidth]{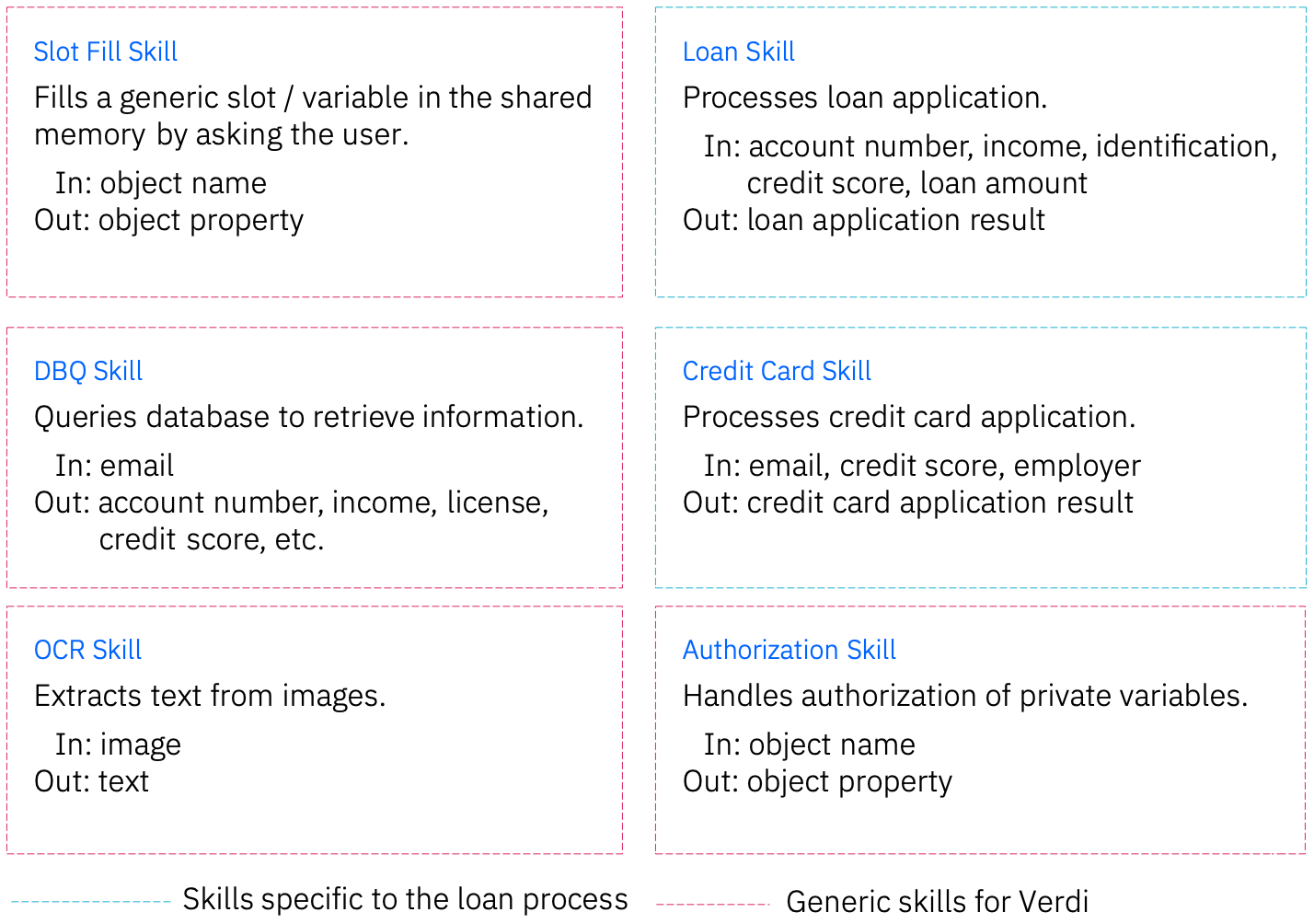}
\caption{The catalog used in Figure \ref{fig:demo}.}
\label{fig:catalog}
\end{figure}

\subsection{Compilation to PDDL}
\label{pddl}

The next step is to convert the agent specification into a planning model. 
Given our problem is closely related to fully observable non-deterministic planning (FOND) \cite{cimatti2003weak}, 
we decided to map the agent specification into a non-deterministic 
planning model with additional considerations provided to support 
problem elements that are uncharacteristic of standard 
non-deterministic planning problems. 
More on this later in Section \ref{subsec:exe}.

First, we create a type for each element that appeared 
in the specification and a type for each individual 
input-output pairs of the skill function specification. 
We create an object for each element type 
% (which for the example domain  was limited to one element per type) 
and one object each for the pairs. 
From the shared ontology, we also incorporate the subsumption information 
into the planning model by converting it into a type hierarchy
(e.g. a pie chart is a type of plot -- so if a skill can return
a pie chart it can also try to respond to a request for a plot). 
In order to use popular off-the-shelf planners 
this means restricting the relations 
to a tree-like hierarchy. 
% We can do this by converting hierarchy into a tree by introducing new types or updating the grounder to allow for non-tree like type hierarchies. Going forward, we can also allow the incorporation of other types of information from the ontology similar to \cite{jeorg}. 

We are interested in establishing the value of different variables in
the shared memory so that the relevant
skills and agents can be invoked with the requisite knowledge required
for their operation (as per their specification). 
Thus, the central fluent of the planning problem is $\textsf{(known ?x)}$. 
The planner itself does not care about the specific value 
that each variable takes but rather reasons with the signature
of each variable, i.e. whether their value is known or not, or 
whether it was attempted and cannot be known.

In terms of action specification, we made use of an all outcome 
determinization \cite{yoon2007ff}
of the non-deterministic model to plan with. 
For each skill $\phi$, whose functional specification can be represented by the tuple $\phi = \langle ([[i, \{o\}_i], \ldots])\rangle$, we create a different action for each possible input and output pair. So just for the tuple $(i, \{o_i^1,.., o_i^k\})$ we create $k$ possible lifted actions each meant to capture the ability of the skill to achieve the specific outcome. For each action $a_{i,1}^{\phi}$, the precondition enforces that the value of each input element in $i$ is known and the effects ensure that for each output element in $o_i^1$ the known value corresponding to that element is set to true. All example plans presented in this paper have been generated using the agile version of Cerberus \cite{katz2018cerberus}.

\subsubsection{Internal Skills and Other Constraints}
Unlike the external skills, we allow for more flexibility when compiling internal skills. 
For example, in the example considered later, we have two internal skills. 
A slot filling skill that allowed querying the user for the value of a specific element
(which basically has an add effect that sets that element to be known)
and an authorization skill that authorizes the use of certain skills over sensitive information. 
In addition to the specific compilation of inputs and outputs, we also allow the inclusion of 
more global constraints. One of the examples we had was this need to perform authorize action 
before sending sensitive information (listed in the specification) to skills. We did this by 
adding an authorize precondition to skills that took sensitive information as an input.

\subsection{Goal Reasoning Engine}
\label{goals:now}

The final component of the planner-based orchestrator is the goal:
the sequences composed will be trying to achieve this goal.
A goal can come from many sources:

\begin{itemize}
\item[-]
The most common goal is derived from the user utterance. 
Every time a user utterance is received,
the goal reasoning engine analyzes it and ascribes a symbolic goal to it. 
The example in Section \ref{example} is of this type. 
In most cases, is just to set the value of an element to known, but in general,
it is a partial state that may be a conjunction that calls for the value of 
several such elements to be determined  -- e.g. a user goal to take a holiday
can resolve to a requirement that a hotel and a flight are booked. 
\item[-]
We mentioned before that an user utterance is just one form of an event
in the system. There could be other events that trigger a goal -- e.g.
a user in a DevOps domain can ask the assistant for a web service 
to be brought up and this can be a 
maintenance goal. The planner will be triggered in this scenario
when an alert pops up notifying that the service has gone down.
\item[-]
Goals may be implicit as well -- e.g. in Algorithm \ref{s3}
the proposed orchestrator can take the Sequencer's role. Then the 
implicit goal is just to make sure that the set of agents 
returned by the Selector are sequenced correctly. 
\end{itemize}

Usually, the mapping from user utterance to goals 
is a domain-specific one and needs to be specified for each agent. 
The initial state is provided by the current set of known values for each element.
This completes the compilation of the orchestration problem to a 
planning problem.

We maintain the current set of goals as a stack. Every request 
from the user gets pushed onto the stack as a new goal. 
At any given time the assistant focuses on achieving the goal 
on top of the stack and whenever the current goal is completed (or 
stopped as per the user's request), 
the assistant falls back to the previous goal on the stack. 
Each time it moves to the previous goal, the user is 
notified about the new goal being pursued, giving 
them an opportunity to either stop or add a new goal.
The goal stack can also be used to augment an interaction with new goals: e.g.
the assistant can prompt the user for a new credit card when a loan application
is completed. Such goal extensions need to be specified by the developer but 
can also be learned from statistical patterns over repeated interactions 
with end users.

% For cases where there is a single object per element, the goal is tied to that object. 
% For cases where we may use multiple objects per type (to allow duplicate objects of the same type), we could use a disjunction in the goal. 

% The goal of the planning problem becomes to set the value of the element known. 
% For cases where there is a single object per element, the goal is tied to that object. 
% For cases where we may use multiple objects per type (to allow duplicate objects of the same type), we could use a disjunction in the goal. Usually, the mapping from user utterance to goals is a domain-specific one and needs to be specified for each agent. 
% The initial state is provided by the current set of known values for each element.

\subsection{Learning Through Execution}
\label{subsec:exe}

As mentioned earlier, the setting we are looking at is not the standard non-deterministic 
planning setting studied in literature. 
In fact, many of the non-deterministic aspects of skill execution could be tied to 
unobservable factors and incomplete specifications. 
In such cases, many of the standard assumptions made for non-deterministic planning 
like fairness \cite{sardina2015towards} are no longer met
\cite{srivastava2016metaphysics,illanes2019generalized}. 
Thus we cannot directly apply traditional non-deterministic planning but 
we will use the information gathered through observation of execution to 
overcome such limitations. Currently, we allow for two such mechanisms:

\subsubsection{Loops.}
To prevent the system from getting trapped in loops trying to establish specific 
values, we introduce a new predicate called \textsf{cannot\_establish} that takes in two arguments: the skill (or more specifically the skill input-output pair) 
and the element in question. 
For each output of an action, we add a negation of \textsf{cannot\_establish} for the corresponding output element and the skill pair corresponding to that skill 
into the precondition. 
This means that if the value for the predicate is set true in the initial state
then the action would not be executed. 
This value is established by the central execution manager, 
that monitors all the skills that have been executed till now and maintains 
a counter of how many times a skill has been unsuccessfully 
called to establish some value. 
When the counter crosses the limit set in the specification, 
the corresponding value is set in the initial state for all future planner calls.

\subsubsection{Model Refinement.}
The next problem we wanted to address is that
these specifications are inherently incomplete. 
Thus the planner may try to call the skills using values that may not be 
valid or expect outputs for certain inputs that the skill may not be able 
to generate:
e.g. not all variables can be slot-filled through conversation with the user. 
We expect currently that such situations can be detected by either the skill 
itself or the execution manager and once such an incorrect invocation of the
skill is detected the corresponding grounded action is pruned out of the model.

\begin{figure}[tbp!]
\centering
\includegraphics[width=0.85\columnwidth]{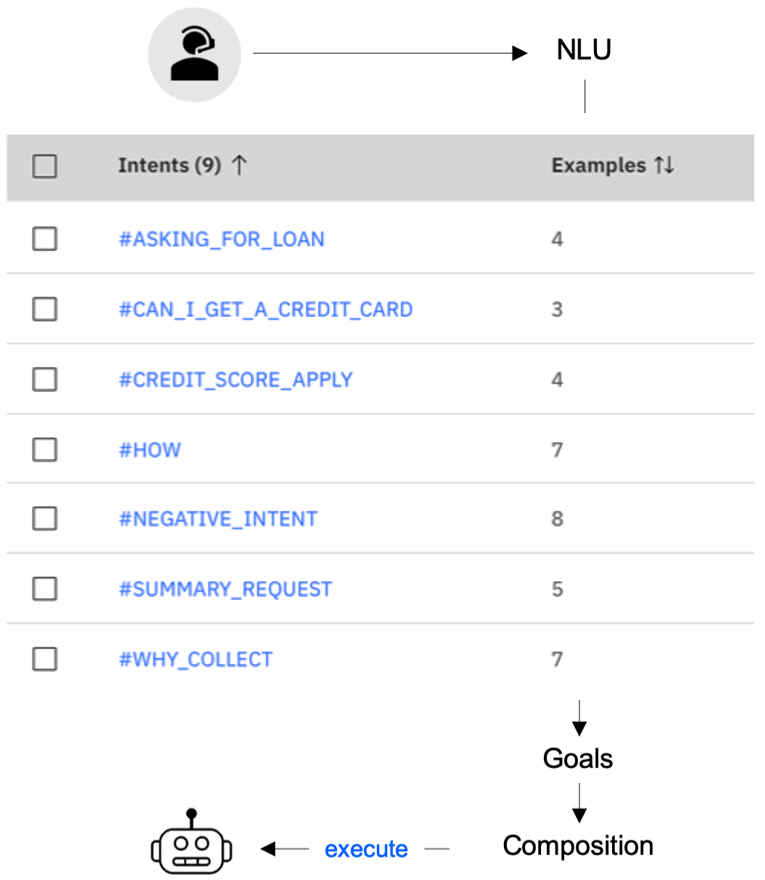}
\caption{The entity recognizer inside the goal reasoning engine.
Some intents (e.g. loan or credit card) correspond to user goals 
while others (how or why questions) are used to manage
domain-independent conversation patterns.}
\label{fig:nlu}
\end{figure}

\begin{figure*}[!th]
\includegraphics[width=0.9\textwidth]{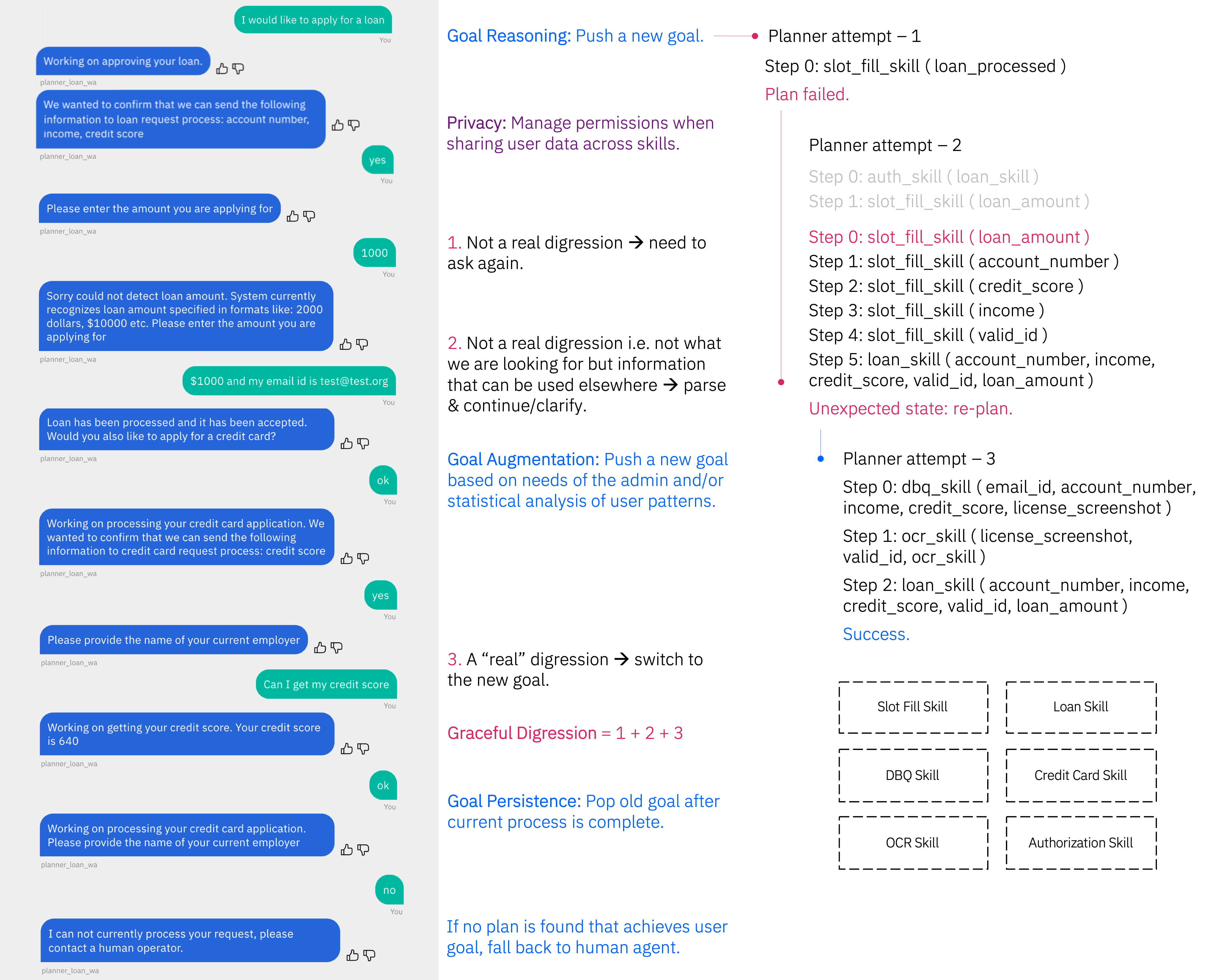}
\caption{Detailed example illustrating stateful 
orchestration of skills in the assistant to achieve
evolving user goals in the banking domain.
A sample conversation with the assistant
is embedded on the left, the middle
provides highlights of interesting stateful
orchestration patterns, and finally,  
the right illustrates the evolving
plans in the backend to support this conversation.}
\label{fig:demo}
\end{figure*}

\subsection{Illustrative Example}
\label{example}

We use a sample banking scenario to illustrate key properties 
of the composed aggregated assistant. 
The assistant here is meant to help users in a banking domain with
submitting a loan and credit card applications, 
which includes gathering relevant information from the users and 
submitting them to the appropriate channels and reporting the results. 

The agent has access to a catalog of skills (Figure \ref{fig:catalog}) including
ones that submit the loan or credit card applications, a skill to
perform OCR, a skill to retrieve information from a database (from where it can
retrieve records by providing the user's email id or account number). 
The agent also has access to two Verdi internal skills: the slot filling skill 
gathers information from the user for variables in the shared 
memory\footnote{The implementation of the slot filling skill
determines which slots can be filled.
This is not part of the specification and thus not known to the planner:
it gets to know when it executes.
}
and an authorization skill meant to confirm with users 
whether to send sensitive information to the skills. 

The goal reasoning engine includes an NLU (Natural Language Understanding) component
(Figure \ref{fig:nlu}) capable of recognizing intents, 
including asking for a loan, a credit card, 
requesting to retrieve and display information, 
asking for explanations, summaries, and to stop the current request,
and so on. 
Each intent is translated to their symbolic goals 
and managed by the goal reasoning engine during execution.

% We will describe in the next section in a bit more detail
% how a goal stack manages these behaviors.

% \subsection{Properties of Stateful Orchestration}

Figure \ref{fig:demo} provides step by step details of the 
orchestration patterns and the plans created and discarded
in the background. A few salient features below:

\subsubsection{Goal Seeking Behavior}

Clearly noticeably is the goal seeking behavior of the orchestrator.
At any moment it tries to create the shortest sequences of agents
or skills for its catalog so as to achieve a goal. This is the 
primary reason to use a planner-based orchestrator and it motivates
a new class of aggregate assistants that can move beyond the
episodic or one-shot interactions afforded by the state of the art. 

\subsubsection{Graceful Digression}

One powerful feature of having an orchestrator than maintains state
is to be able to digress gracefully. This involves: 
1) Asking again if something is not clear;
2) Parsing additional or seemingly unrelated 
information for other goals; and
3) Switching to a new goal if necessary and switching 
back to the old one, once done.
Modeling digression patterns well is essential to conversational assistants 
and we posit that using a stateful orchestrator is the only way to induce 
such behavior. Graceful digression patterns is a direct 
consequence of the goal stack.

\subsubsection{Authentication \& Privacy}

Finally, one concern in the composition of an aggregated assistant
is that of the information that will eventually get distributed across
the various agents, skills, and services plugged into it. 
We considered two privacy models: 
one where every skill or agent is authenticated once 
%at the time of sharing 
the first time any private information is shared with it
and one where each data
element is authenticated once at the time of fetching. 
The former requires less back and forth with the user and so 
we went with this as the default.
% , as we explained towards the
% end of Section \ref{pddl}.

\section{Automated Orchestration and XAIP}

Even though using planners to compose stateful orchestration 
patterns on the fly allowed us to model the complex 
conversational patterns above, from the end-user 
perspective, we have a problem: in the example in
Figure \ref{fig:demo} we noted the internal processes
going on in response to the user utterances but
none of this is actually visible to the user. 
This means when the goal is achieved, a series
of events have unfolded in the background for the assistant 
to reach that conclusion unbeknownst to the user. 
As such, this can be quite unsettling, as seen
in the sudden response with the loan result 
after a couple of interactions with the assistant 
acquiring information from the user.
We propose to navigate this situation
of the automated orchestration approach using
what'', ``how'', and ''why'' questions.
There are many reasons to surface the 
internal processes to the end-user on demand:

\begin{itemize}
\item[-] 
The transparency of automated compositions for the user 
of the assistant who cannot see the plans being made and discarded 
in the background;
\item[-]
Above, where the end-user is a subject matter expert who can 
use the explanations to affect better compositions;
\item[-]
Above, where the end user has a right to an explanation, 
as per GDPR guidelines \cite{kaminski2019right}. 
Even though such guidelines may not apply to the constraints of 
the underlying process \cite{wachter2017right}, 
they do certainly apply to data subjects of which the user is 
one, as is evident in the examples provided where multiple skills
and agents fetch and use user information in the background 
to achieve user or system goals. 
\end{itemize}
% \todo{The overall flow of the XAIP components of the system is presented in Algorithm \ref{algo}}.

\subsection{``What?'' Questions}

The first step towards achieving transparency is to be 
able to provide summaries \cite{amir2018highlights,sreedharan2020tldr}
of what was done in the backend back to the user. 
Instead of simply dumping the entire steps of the plan, 
our focus here was to provide only the most-pertinent 
set of information to the user and provide them with 
ability to drill down as required. 
This is motivated not just by the fact that the plan may have many steps, 
but also by the online nature of the planning problem -- the agent may have 
performed steps that were in the end not pertinent to the final solution. 
To provide information central to achieving the goal, 
we extract the landmarks corresponding to the goal from the original initial 
state, since regardless of the exact steps all information corresponding to 
landmark facts should have been collected to get to the goal. 

\begin{figure}[tbp!]
\centering
\includegraphics[width=0.9\columnwidth]{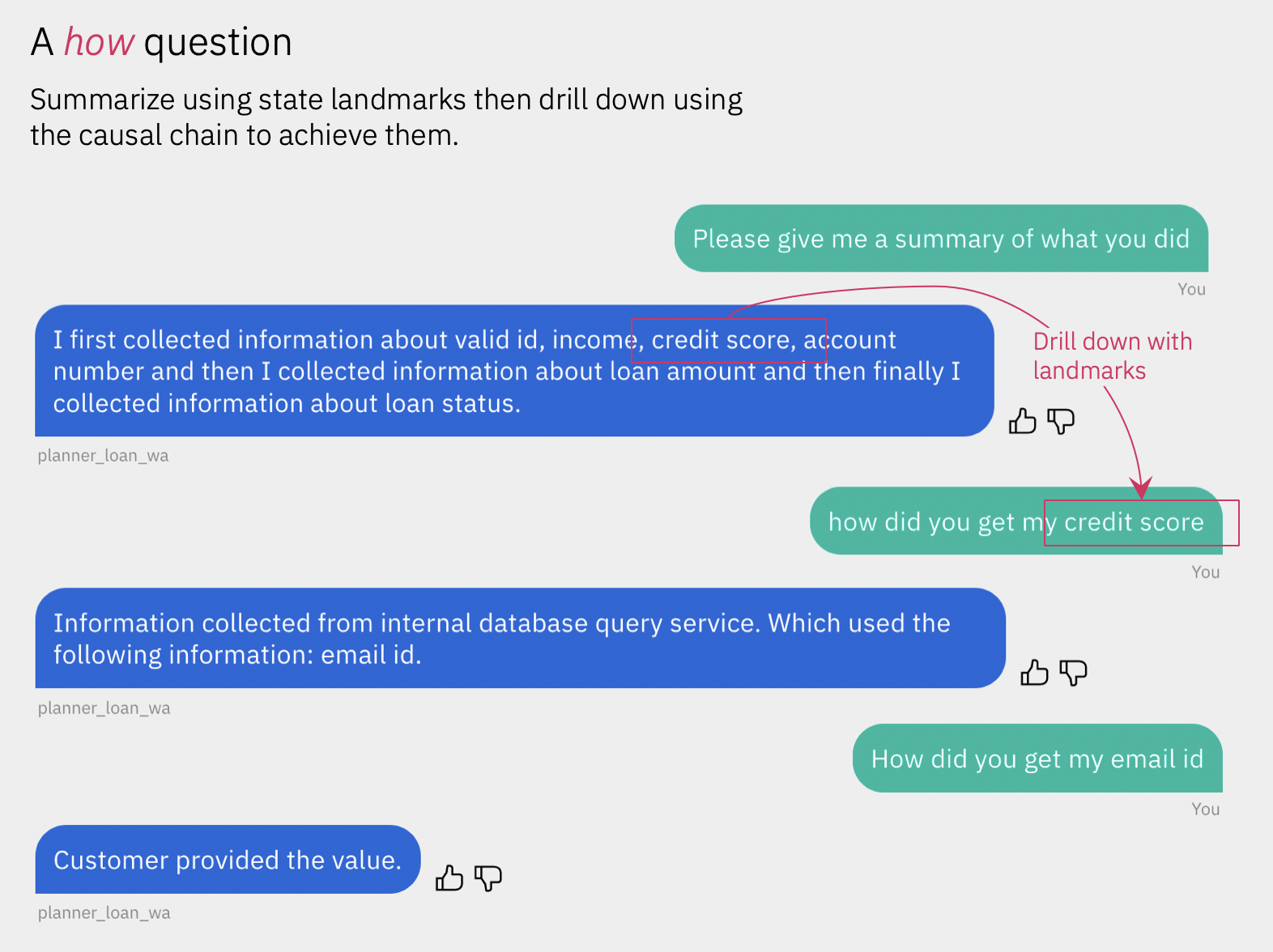}
\caption{
Example of a summary followed by the 
user drilling down with "how?" questions.
}
\label{fig:how}
\end{figure}

Such techniques have been used recently to summarize policies 
\cite{sreedharan2020tldr}. Unlike those cases, the system here 
is also learning about the domain during execution.
So we also need to incorporate any \textsf{cannot\_establish} facts 
that have been learned along the way into the initial state 
and also remove any actions that may have been pruned along the way. 
This means even though the first time the planner was called, 
it believed it could establish the goal with a single step, 
through the execution it realized it needs to gather multiple information items.
Figure \ref{fig:how} shows the summary generated for the current 
demo domain.\footnote{This 
currently uses a topological sort of landmark ordering 
and then filters out the compilation specific variable. 
Moving forward, we plan to use the actual order in which 
data was collected.
} The actual summary text presented to the user uses a template-based generation,
that relies on ordering between the landmarks generated via a topological
ordering of landmarks corresponding to the establishment of information (while ignoring all compilation specific fluents). For the scenario reported in the paper, we focused on fact landmarks generated using methods discussed in \cite{keyder2010sound}.

% In the future we could also go with just using the order it was collected in the executed plan.

% Figure \ref{fig:why}, shows the example explanation provided for our demo domain.

\subsection{``How?'' Questions}

Once a summary is provided, we allow the user to drill
down on any aspect of it by asking how that was achieved.
This involves identifying the agent or skill that established 
the value of the element and providing that detail to the user. 
Here we rely on the simple description provided by the skill author in the 
specification. In addition to showing the skill, we also provide the inputs 
that the skill took, so
that if the user requires they can chain even further back.

\subsection{``Why?'' Questions}

Furthermore, the user can also go forward from that 
fact and explore why that was required to be established,
in terms of its role in achieving the goal.  
We adapt the popular idea of using the causal chain 
from an action to the goal as the justification for its role in the plan (c.f. \cite{veloso1992learning,seegebarth2012making,bercher2014plan,ai-comm,kambhampati1990classification}) to individual facts. 
We identify whether an action contributed something to the goal by performing a regression from the goal over each action that was executed by the agent. 
For a skill execution that took as input the set of facts $i$ and generated an output set $o_i^j$, for a current regressed state $s_i$, the next state $s_{i-1}$ is provided as $(s_i \setminus o_i^j) \cup i$. 
Now we can allow for the fact that the system may have made missteps, by ignoring any actions that do not contribute something to the current regressed state. 
We stop as soon as we reach an agent or skill that contributes to the regressed state and used the fact in question. We can either provide the full causal chain to the goal (by build such chains on the fly from the goal during the regression process) or just provide this final action. 
% \todo{Figure \ref{fig:why} shows an example ``Why'' explanation for the scenario considered.}

\begin{figure}[tbp!]
\centering
\includegraphics[width=0.9\columnwidth]{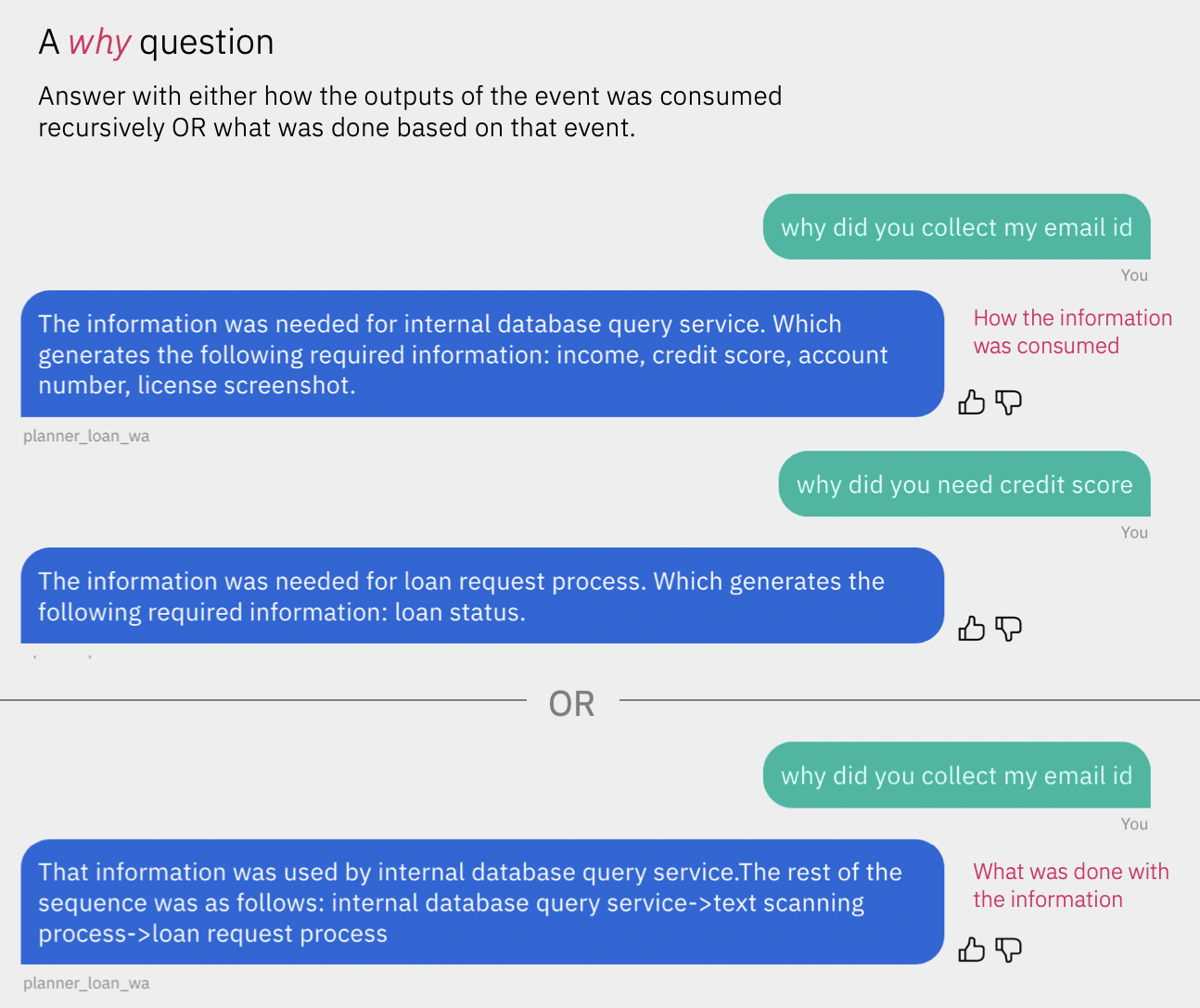}
\caption{Example of two ways to answer "why?" questions.}
\label{fig:why}
\end{figure}

\begin{figure}
\centering
\includegraphics[width=0.9\columnwidth]{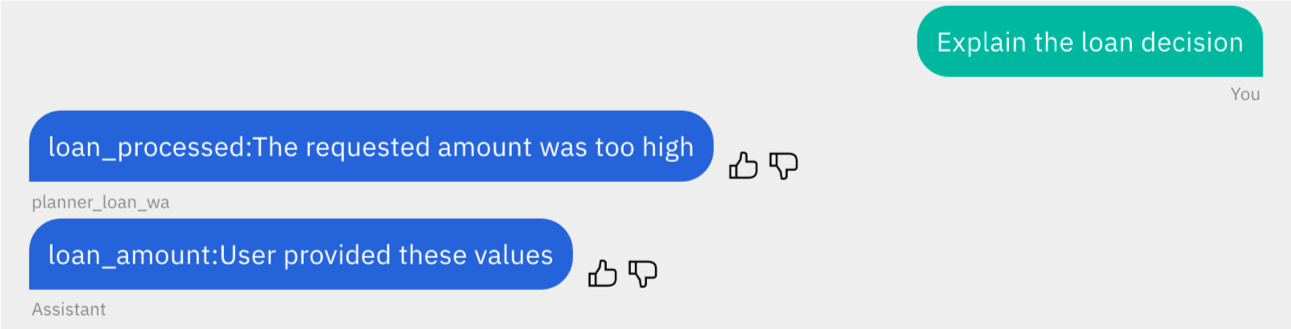}
\caption{Example of augmenting the explanation of the 
orchestration process with that of a decision made
by a skill.}
\label{fig:wip}
\end{figure}

\subsection{Work in Progress: Explanation Chaining}

The above discussions were focused on generating explanations for 
the orchestration patterns only but do not try to explain why the individual
agents and skills made the decisions they made.
They may be using AI components to generate their decisions and we may 
want the explanations for these individual decisions to be provided by the agent or skill itself. 
Thus, we envision a third component of an agent (c.f. Section \ref{definitions}):
$\Phi.explain$ which provides an explanation of its output in terms of 
its inputs, much like feature attribution explanations  \cite{sundararajan2019many}. 

For example, the loan approval skill might say your credit score is too low. 
The system can provide assistance here is by tracing the source of such information: e.g. it can go back and identify where it
found the credit score that was passed to loan approval skill. 
This creates a chain of skills, 
with the explanation for the output of one agent skill 
being clarified with an explanation from another. 
Currently, the chain is terminated, when the backward chaining reaches 
an information that is known to the user
(such as one specifically provided by them). The example explanation for the scenario is presented in Figure \ref{fig:wip}.
For now, we limit chaining to the feature with the 
highest weight as attributed by the explanation.

\section{Conclusion and Future Work}

In this paper, we introduced a new type of (conversational) 
assistant that is becoming increasingly popular: an 
``aggregated assistant'' realised as an orchestrated 
composition of skills and agents. 
We looked at the role of automated planning in it, in how 
it can facilitate the automated composition of such assistants, and
this can lead to loss of transparency. 
We showed how techniques from the XAIP community in surfacing 
causal information of planning domain to end users can help 
in mitigating such concerns of transparency in automatically
constructed aggregate assistants. 
In the following, we expand on the presented ideas
with some interesting avenues of future work.

\subsubsection{What is a goal?}

In the approach described in the paper, we only talked about 
``logical'' goals derived from the user utterance. This does not 
always have to be a case. A more general form of a goal
are ``metric'' goals -- these are observed metrics 
of the processes being modeled by the assistant.
In the planning parlance, these can often be modeled 
as numeric fluents \cite{fox2003pddl2}
as properties of an individual skills or agents:
e.g. their health, accuracy, money cost, etc.
But not always. Cases where 
it is difficult to model metric goals is when
they are properties of the entire process and 
not individual actions in the domain -- this 
becomes harder when such properties are ill defined.
Examples include customer satisfaction, timespan of 
processes (when they are not modeled), overall 
expertise requirement of a process, and so on. 
In the context of business processes, these are often
referred to as Key Performance Metrics \cite{parmenter2015key}
that the system admin monitors and cares about, and eventually
we would want to compose assistants that cater to the desired
``KPI goals'' of the composed assistant: e.g. ``achieve
compositions with target metric $\mathcal{M}$''
based on historical data. 

Another avenue of extension here is in the way 
the logical goals are extracted from the user utterance.
Currently, as we outlined in Section \ref{goals:now},
we use intent classifiers coded in by the developer of the
assistant. This, again, requires modeling overhead. 
There are two possible ways to get around this.
One can match the user utterance to the specification 
and natural language description of a skill or agent in the catalog
and use (the use of) that component as the end goal.
This requires less modeling overhead but is also
less expressive since, in general, goals are attributed
to a state of the system which may not correspond to the effects of
a specific skill: e.g. a goal to book a trip would require
hotel, flight, etc. The other way to slowly generate this
mapping is to use a ``cold starting'' scheme, as described below.

\begin{algorithm}[!th]
\SetAlgoLined
\DontPrintSemicolon
$\mathcal{G}\leftarrow \textrm{ Goal set}$\\
$\mathcal{H} \leftarrow \textrm{Execution History}$
\tcp*{A sequence of tuples $\langle i, \phi, o_i\rangle$ ending at the goal}
% \tcp*{A sequence of tuples of the form $\langle i, \phi, o_i\rangle$ 
% and ends with goal achievement}
\SetKwFunction{Fwhy}{Why-next-action}
\SetKwFunction{Fhow}{How-last-action}
\SetKwFunction{Fchain}{Chain-explanation}
\SetKwFunction{Flandmark}{Landmark-Extractor}
\SetKwFunction{Fgoal}{Derive-Goal}

\vspace{5pt}
\While {$\mathcal{E}$}{
$\mathcal{G}, f$ $\leftarrow$ \FGoal$(\mathcal{E})$\\
\If {$\mathcal{G} \rightarrow$ ``Why''}{
\Fwhy($f$)
}
\ElseIf {$\mathcal{G} \rightarrow$ ``How''}{
\Fhow($f$)
}
\ElseIf {$\mathcal{G} \rightarrow$ ``What''}{
\Flandmark($f$)
}
}

\vspace{5pt}
\SetKwProg{Fn}{Function}{:}{}
\Fn{\Fhow{$f$}}{
\For{$\langle i, \phi, o_i\rangle \in \mathcal{H}$}{
\If{ $f \in o_i$}{
\Return $(i, \phi)$
}
}
}

\vspace{5pt}
\SetKwProg{Fn}{Function}{:}{}
\Fn{\Fwhy{$f$}}{
$\mathcal{R}\leftarrow\mathcal{G}$\\
\For{$\langle i, \phi, o_i\rangle \in \textsf{Reverse}(\mathcal{H})$}{
\If {$\mathcal{R} \subseteq o_i$}{
$\mathcal{R} =(\mathcal{R} \setminus o_i) \cup i$ 
}
\If{ $f \in i$}{
\Return $\phi$
}
}
}

\vspace{5pt}
\SetKwProg{Fn}{Function}{:}{}
\Fn{\Fchain{$f$}}{
\While{$f$ is not provided by user}{
$i, \phi \leftarrow\ $\Fhow(f)\\
$\phi.\textrm{explain(f)}$
}}

\caption{Explanation Flow of Control}
\label{algo:overall}

\end{algorithm}

\subsubsection{Cold starting the orchestrator.}

One of the topics we discussed in Section \ref{subsec:exe}
was the refinement of the skill and agent specifications 
over time by learning from execution patterns. 
This can help out especially in reducing the modeling 
overhead on the developer or the admin of the assistant. 
One can start off with a minimal specification and an
S3-orchestrator, observe a few interactions
(or simulate the assistant) and 
and refine those specifications over time, before 
activating the planner-based orchestrator. 
The latter buys us the more complex stateful orchestration
patterns as we discussed; and this ``cold start'' scheme
reduces the need for detailed or complete specifications
up front. There are many existing approaches to learning
PDDL models from traces \cite{gil1994learning,zhuo2013refining}; 
including ones that extend PDDL \cite{nguyen2017robust} 
to handle the incompleteness of the learnt model
in terms of {\em possible} conditions.

\subsubsection{Agent Preview.}

Finally, you may have noticed, that we never used the 
agent preview in the planning-based orchestrator. 
That is because the planner chose a composition 
and only replanned at the time of execution: this flow 
meant that, as comapred to the S3-orchestrator, the 
planning-based orchestrator had no separate selector
and sequencer. That is just the job of the planning problem. 

However, this flow also means that there will be frequent
calls to replan when the execution fails; and this will
affect the assistant's time to response to the end user. 
The way to mitigate this would be to use the preview 
in the planning process itself to compute higher fidelity
plans. This is an optimization in the backend and not immediately
visible to the end user (in terms of quality of compositions)
unless it affects the latency. Interestingly, this can also help the 
orchestrator to deal with agents or skills that may change state
and thus should not be executed blindly, and even to hand-off
some of the specification overhead to the implementation 
of the agent or skill itself so that they can respond with
possible outcomes of an execution call in the preview call.

% \clearpage
% \small
\bibliographystyle{aaai}
\bibliography{bib}

\end{document}